\documentclass{article}
\usepackage{fullpage}
\usepackage{amsmath}

\usepackage{times}
\usepackage{graphicx} % more modern
\usepackage{subfig}
\graphicspath{{./figs/}}

\usepackage[hidelinks]{hyperref}
\usepackage{fancyvrb}
\usepackage{authblk}
\makeatletter
\newcommand{\eqnum}{\leavevmode\hfill\refstepcounter{equation}\textup{\tagform@{\theequation}}}
\makeatother

\title{Interactive Music Generation  with Positional Constraints using Anticipation-RNNs}

\author[1, 2]{Ga\"etan Hadjeres}
\author[3, 4]{Frank Nielsen}

 \affil[1]{LIP6, Universit\'e Pierre et Marie Curie}
 \affil[2]{Sony CSL, Paris}
 \affil[3]{\'Ecole Polytechnique, Palaiseau, France}
 \affil[4]{Sony CSL, Tokyo}

\date{}
\begin{document}

\maketitle

%break url after -
\expandafter\def\expandafter\UrlBreaks\expandafter{\UrlBreaks%  save the current one
  \do\-}

\begin{abstract}
  Recurrent Neural Networks (RNNS) are now widely used on sequence
  generation tasks due to their ability to learn long-range
  dependencies and to generate sequences of arbitrary length. However,
  their left-to-right generation procedure only allows a limited
  control from a potential user which makes them unsuitable for
  interactive and creative usages such as interactive music generation.

  This paper introduces a novel architecture called
  \emph{Anticipation-RNN} which possesses the assets of the RNN-based
  generative models while allowing to enforce user-defined positional
  constraints.  We demonstrate its efficiency on the task of generating melodies
  satisfying positional constraints in the style of the soprano parts
  of the \emph{J.S. Bach chorale harmonizations}.

  Sampling using the
  Anticipation-RNN is of the same order of complexity than sampling
  from the traditional RNN model. This fast and interactive
  generation of musical sequences opens ways to devise real-time
  systems that could be used for creative purposes.
\end{abstract}

\section{Introduction}
\label{sec:introduction}
Recently, a number of powerful generative models on symbolic music have been
proposed. If they now perform well on a variety of different musical
datasets, from monophonic folk-music
\cite{sturm:celtic:melody:arxiv:2016} to polyphonic Bach chorales
\cite{liang2016bachbot}, these models tend to face similar
limitations: they do not provide musically-interesting ways for a user
to interact with them. Most of the time, only an input seed can be
specified in order to condition the model upon: once the generation is
finished, the user can only accept the result or regenerate another
musical content.  We believe that this restriction hinders creativity
since the user do not play an active part in the music creation process.

Generation in these generative models is often performed from left to
right; Recurrent Neural Networks (RNNs) \cite{Goodfellow-et-al-2016}
are generally used to estimate the probability of generating the next
musical event, and generation is done by iteratively sampling one
musical event after another. This left-to-right modeling seems natural
since music unfolds through time and this holds both for monophonic
\cite{sturm:celtic:melody:arxiv:2016,Colombo2017} and polyphonic
\cite{boulanger2012modeling,liang2016bachbot} music generation tasks.
However, this does not match real compositional principles since
composition is mostly done in an iterative and non-sequential way
\cite{Balmer699}. As a simple example, one may want to generate a
melody that ends on a specific note, but generating such melodies
while staying in the learned style (the melodies are sampled with the
correct probabilities) is in general a non trivial problem when
generation is performed from left to right. This problem has been
solved when the generative model is a Markov model
\cite{pachet2014imitative,Papadopoulos2015} but remains hard when
considering arbitrary RNNs.

In order to solve issues raised by the left-to-right sampling scheme,
approaches based on MCMC methods have been proposed, in the context of
monophonic sequences with shallow models \cite{Sakellariou2017} or on
polyphonic musical pieces using deeper models
\cite{hadjeres2016style,DBLP:conf/icml/HadjeresPN17}. If these MCMC
methods allow to generate musically-convincing sequences while
enforcing many user-defined constraints, the generation process is
generally order of magnitudes longer than the simpler left-to-right
generation scheme. This can prevent for instance using these models in
real-time settings.

The problem of generating sequences while enforcing user-defined
constraints is rarely considered in the machine learning literature
but it is of crucial importance when devising interactive generative
models.  In this paper, we propose a neural network architecture
called \emph{Anticipation-RNN} which is capable of generating in the
style learned from a database while enforcing user-defined positional
constraints. This architecture is very general and works with any RNN
implementation. Furthermore, the generation process is fast as it only
requires two function calls per musical event. In
Sect.~\ref{sec:statement-problem}, we precisely state the problem we
consider and Sect.~\ref{sec:model} describes the proposed architecture
together with an adapted training procedure. Finally, we demonstrate
experimentally the efficiency of our approach on the dataset of the
\emph{chorale melodies by J.S. Bach} in
Sect.~\ref{sec:experimental-results}. In Sect.~\ref{sec:conclusion},
we discuss about the generality of our approach and about future
developments.

\section{Statement of the problem}
\label{sec:statement-problem}
We consider an i.i.d. dataset $\mathcal{D} := \{s = (s_1, \dots, s_N) \in \mathcal{A}^N\}$ of sequences of tokens $s_t \in \mathcal{A}$ of arbitrary length $N$ over a vocabulary $\mathcal{A}$.
We are interested in probabilistic models  over sequences $p(s)$ such that
\begin{equation}
  \label{eq:2}
 p(s) = \prod_t p(s_t| s_{<t}),
\end{equation}
where
$s_{<t} = (s_1, \dots, s_{t-1})$ for $t>0$ and $\emptyset$ if
$=0$. This means that the generative model $p(s)$ over sequences is
defined using the conditional probabilities $p(s_t| s_{<t})$
only. Generation with this generative model is performed iteratively
by sampling $s_t$ from $p(s_t| s_{<t})$ for $t = 1..N$ where $N$ is arbitrary. Due to their
simplicity and their efficiency, Recurrent Neural Networks (RNNs) are
used to model the conditional probability distributions
$p(s_t| s_{<t})$: they allow to reuse the same neural network
over the different time steps by introducing a hidden state vector in
order to summarize the previous observations we condition on. More precisely, by writing $f$ the RNN, $\textrm{in}_{t}$ its input, $\textrm{out}_{t+1}$ its output and $h_t$ its hidden state at time $t$, we have
\begin{equation}
  \label{eq:1}
  \textrm{out}_{t+1}, h_{t+1} = f(\textrm{in}_t, h_t)
\end{equation}
for all time indices $t$. When $\textrm{in}_t = s_t$, the vector $\textrm{out}_{t+1}$ is used to define $p(s_{t+1}| s_{<t+1})$ for all time indices $t$ without the need to take as an input the entire sequence history  $s_{<t+1}$.

If this approach is successful on many applications, such a model can only be conditioned on the past which prevents some possible creative use for these models: we can easily fix the beginning $s_{<t}$ of a sequence and generate a continuation $s_{\geq t} = (s_t, \dots, s_N)$ but it becomes more intricate to  fix the end $s_{\geq t}$ of a sequence and ask the model to generate a beginning sequence.

We now write $p_\textrm{unconstrained}(s)$ the probability of a sequence $s$ when no constraint is set. For simplicity of notation, we will suppose that we only generate sequences of fixed length $N$ and denote by $\mathcal{S} := \mathcal{A}^N$ the set of all sequences over $\mathcal{A}$.
The aim of this paper is to be able to  enforce a set of positional constraints
\begin{equation}
  \label{eq:3}
  \mathcal{C} = \{(i, c_i)\}_{i \in I},
\end{equation}
where $I$ is the set of constrained time indexes and $c_i \in \mathcal{A}$ the value of the constrained note at time index $i$.
Ideally, we want to sample constrained sequences
\begin{equation}
  \label{eq:4}  
\mathcal{S}_\textrm{constrained}\{s \in \mathcal{S}, \quad s_i = c_i \quad  \forall (i, c_i) \in \mathcal{C}\}
\end{equation}
with the ``correct'' probabilities. This means that, if we denote by  $p_\textrm{constrained}(s)$ the probability of a sequence $s$ in the constrained model:
\begin{itemize}
\item $p_\textrm{constrained}(s) = 0$, \quad  $\forall s \notin \mathcal{S}_\textrm{constrained}$ \eqnum\label{eq:req1}
\item $p_\textrm{constrained}(s) = \frac{1}{\alpha} p_\textrm{unconstrained}(s), \quad  \forall s \in \mathcal{S}$,\quad  with $\alpha := \sum_{s \in \mathcal{S}_\textrm{constrained}}p_\textrm{unconstrained}(s)$. \eqnum\label{eq:req2}
\end{itemize}
To put it in words, the set of constraints $\mathcal{C}$ defines a subset
$\mathcal{S}_\textrm{constrained}$ of $\mathcal{S}$ from which we want to sample from using the probabilities (up to a normalization factor) given by $p_\textrm{unconstrained}$. However, sampling from $\mathcal{S}_\textrm{constrained}$ using the acceptance-rejection sampling method is not efficient due to the arbitrary number of constraints. Exact sampling from $\mathcal{S}_\textrm{constrained}$ is possible when the  conditional probability distributions are modeled using  models such as Markov models but is intractable in general. This problem in the case of Markov models can in fact be exactly solved  when considering more complex constraints on the space of sequences such as imposing the equality or the difference between two sequences symbols $s_i$ and $s_j$. Generalizations of this problem to other types of constraints are discussed in Sect.~\ref{sec:conclusion}.

\begin{figure*}

  \centering
 
      \includegraphics[scale=0.9]{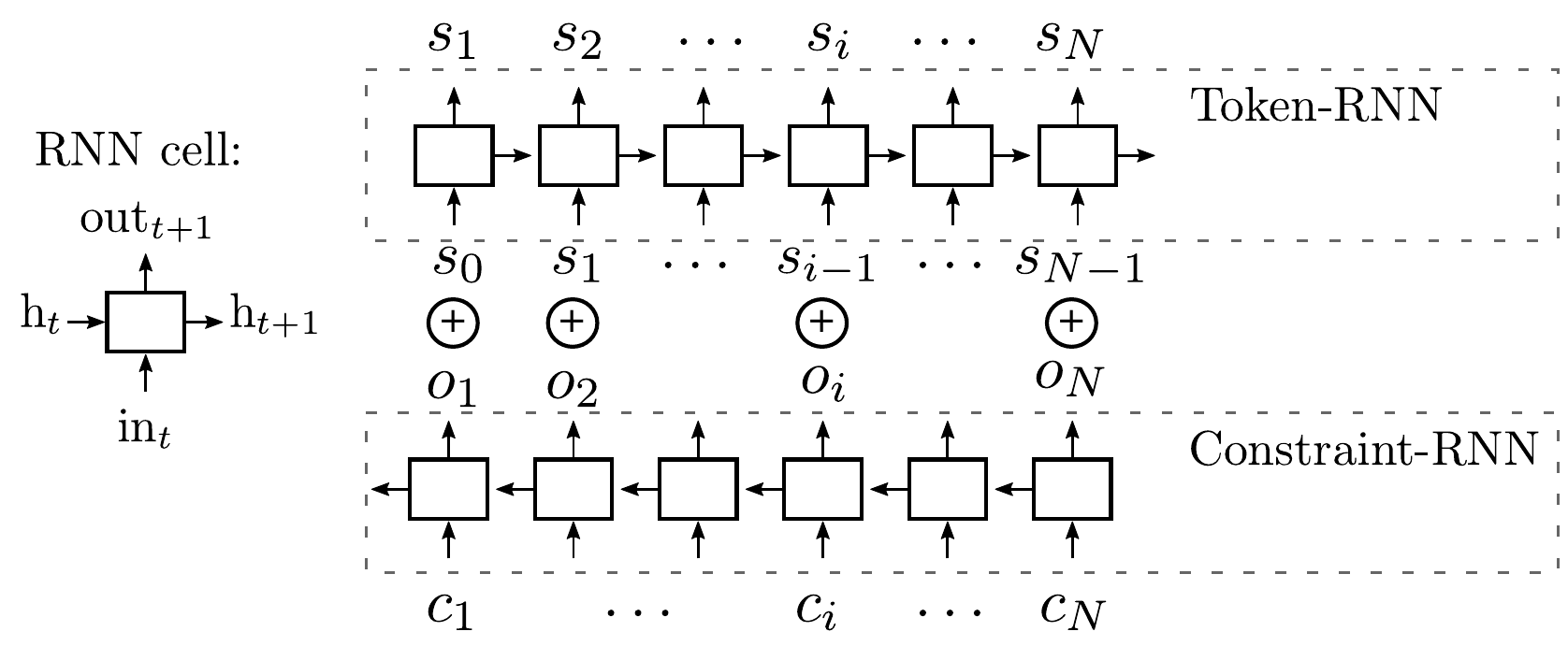}

      \caption[AEs]{Anticipation-RNN architecture. The aim is to predict $(s_1, \dots, s_{N})$
        given $(c_1, \dots, c_N)$ and $(s_0, \dots, s_{N-1})$.}
  \label{fig:archi}
\end{figure*}

\section{The model}
\label{sec:model}
The problem when trying to enforce a constraint $c := (i, c_i)$  is that imposing such a constraint on time index $i$ ``twists'' the conditional probability distributions $p(s_t| s_{<t})$ for $t < i$. However, the direct computation of $p(s_t| s_{<t}, s_i = c_i)$ (using Bayes rule when only $p(s_t| s_{<t})$ is known) is computationally expensive.

The idea to overcome this issue is to introduce a neural network in order to summarize the set of constraints $\mathcal{C}$. To this end, we introduce an additional token $\mathtt{NC}$ (No Constraint) to $\mathcal{A}$ indicating that no positional constraint is set at a given position. By doing this, we can rewrite the set $\mathcal{C}$ as a sequence $c = (c_1, \dots, c_N)$ where $c_i \in \mathcal{A} \cup \{\mathtt{NC}\}$. We then introduce a RNN called \emph{Constraint-RNN} in order to summarize the sequence of all constraints. This RNN goes backward (from $c_N$ to $c_1$) and \emph{all} its outputs are used to condition a second RNN called \emph{Token-RNN}.

This architecture, called \emph{Anticipation-RNN} since the Token-RNN is conditioned on what may come next, is depicted in Fig.~\ref{fig:archi}. We notated by $(o_1, \dots, o_N)$ the output sequence of the Constraint-RNN (for notational simplicity, we reversed the sequence numbering: the first output of the Constraint-RNN is $o_N$ in our notation). The aim of the output vector $o_t$ is to summarize all information about constraints  from time $t$ up to the end of the sequence. This vector is then concatenated to the input $s_{t-1}$ of the Token-RNN at time index $t$  whose aim is to predict $s_t$.

Our approach differs from the approaches using  Markov models in the sense that we directly take into our conditional probability distributions rather than trying to sample sequences in $\mathcal{S}_\textrm{constrained}$ using $p_\textrm{unconstrained}$:
we want our probabilistic model to be able to directly enforce hard constraints.

The Anticipation-RNN thus takes as an input both a sequence of tokens $(s_0, \dots, s_{N-1})$ and a sequence of constraints $(c_1, \dots, c_N)$ and has to predict the shifted sequence $(s_1, \dots, s_{N})$. The only requirement here is that the constraints have to be coherent with the sequence: $c_i = s_i$ if $c_i \neq \mathtt{NC}$. Since we want our model to be able to deal with any positional constraints, we  consider the dataset of couples of token-sequences and constraint-sequences $\mathcal{D}_\textrm{constraint}$ such that 
\begin{equation}
  \label{eq:5}  
\mathcal{D}_\textrm{constraint} := \left\{ \left(s, m(s)\right), \quad \forall s \in \mathcal{D}, \forall m \in \{0,1\}^N\right\},
\end{equation}
where $\{0,1\}^N$ is the set of all binary masks: the sequence of constraints $m(s)$ is then defined as the sequence $(c_1, \dots, c_N)$ where $c_i= s_i$ if $m_i = 1$ and $c_i = \mathtt{NC}$ otherwise.

The sampling procedure is fast since it only needs two RNN passes on the sequence.

\begin{figure}
  \centering
\begin{BVerbatim}
D4 __ E4 __ A4 __ __ __ G4 __ F#4 __ E4 __ __ __
\end{BVerbatim}
  \caption{Melodico-rhythmic encoding of the first bar of the melody of Fig.~\ref{fig:ex}.}
  \label{fig:encoding}
\end{figure}

\begin{figure}
  \centering
  \subfloat[][]{\includegraphics[scale=0.8]{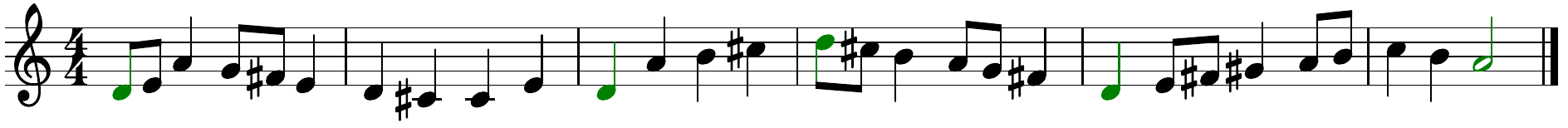}
  \label{fig:ex}}
  
  \subfloat[][]{\includegraphics[scale=0.8]{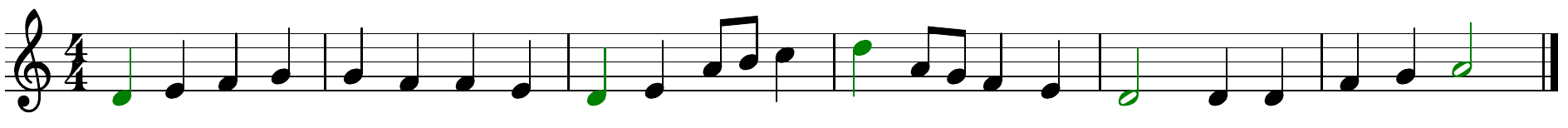}}
  
  \subfloat[][]{\includegraphics[scale=0.8]{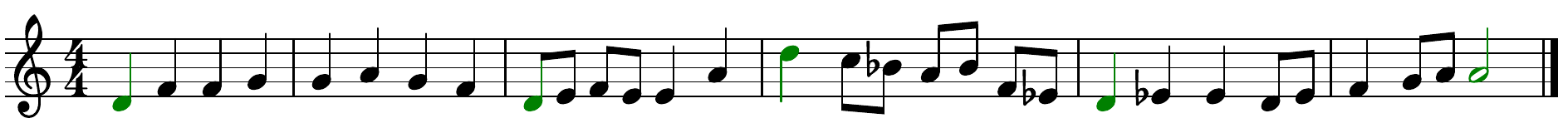}}
    \caption{Examples of generated sequences in the style of the soprano parts of the J.S. Bach chorales. All examples are  subject to the same set of positional constraints  indicated using green notes.
    }
      \label{fig:ex1}
    \end{figure}

    \section{Experimental results}
    
 \begin{figure}
   \centering
   \subfloat[][Constrained case: $p = p_{\text{constrained}}$]{\includegraphics[scale=0.8]{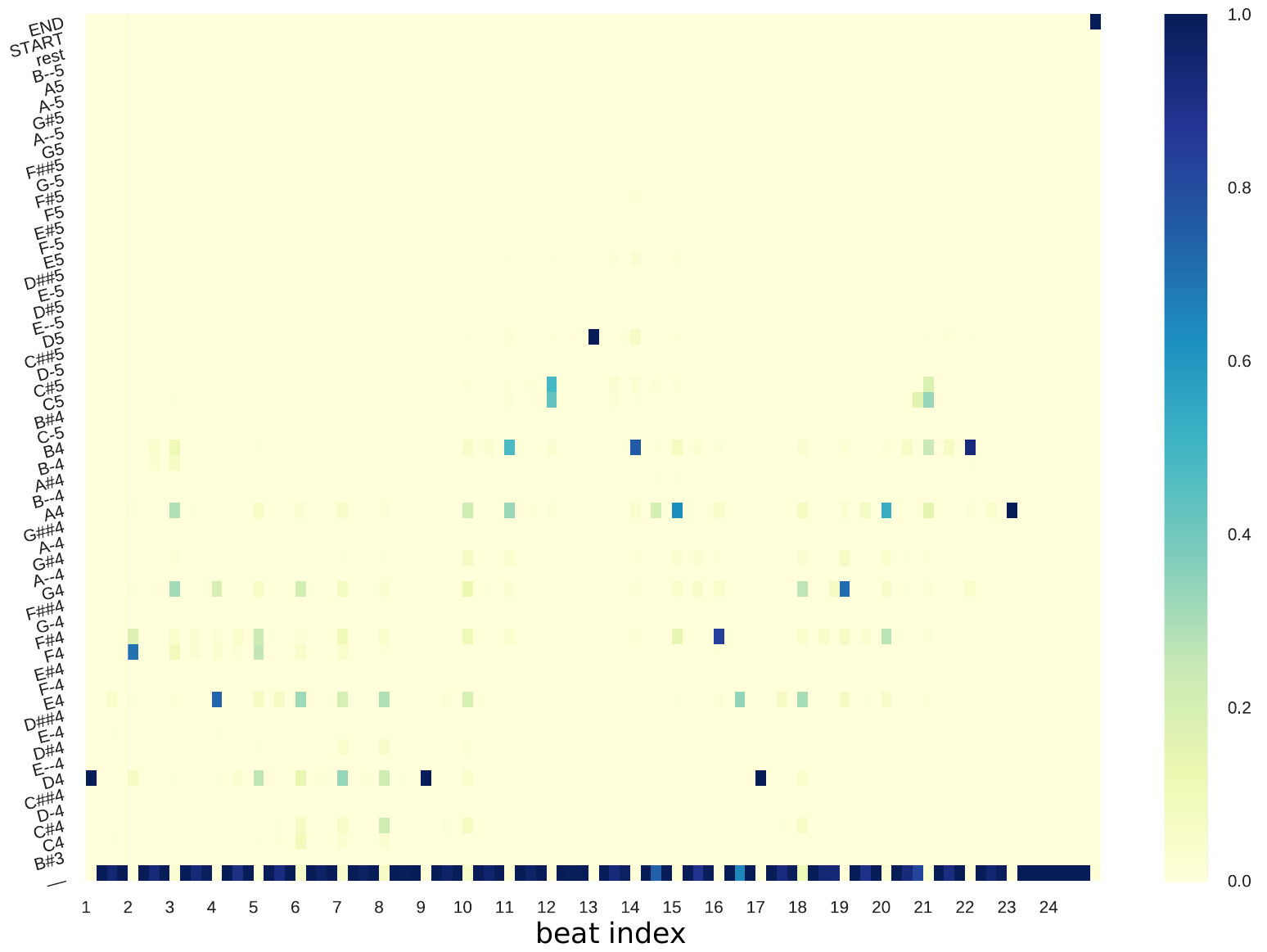}} 
  
   \subfloat[][Unconstrained case: $p = p_{\text{unconstrained}}$]{\includegraphics[scale=0.8]{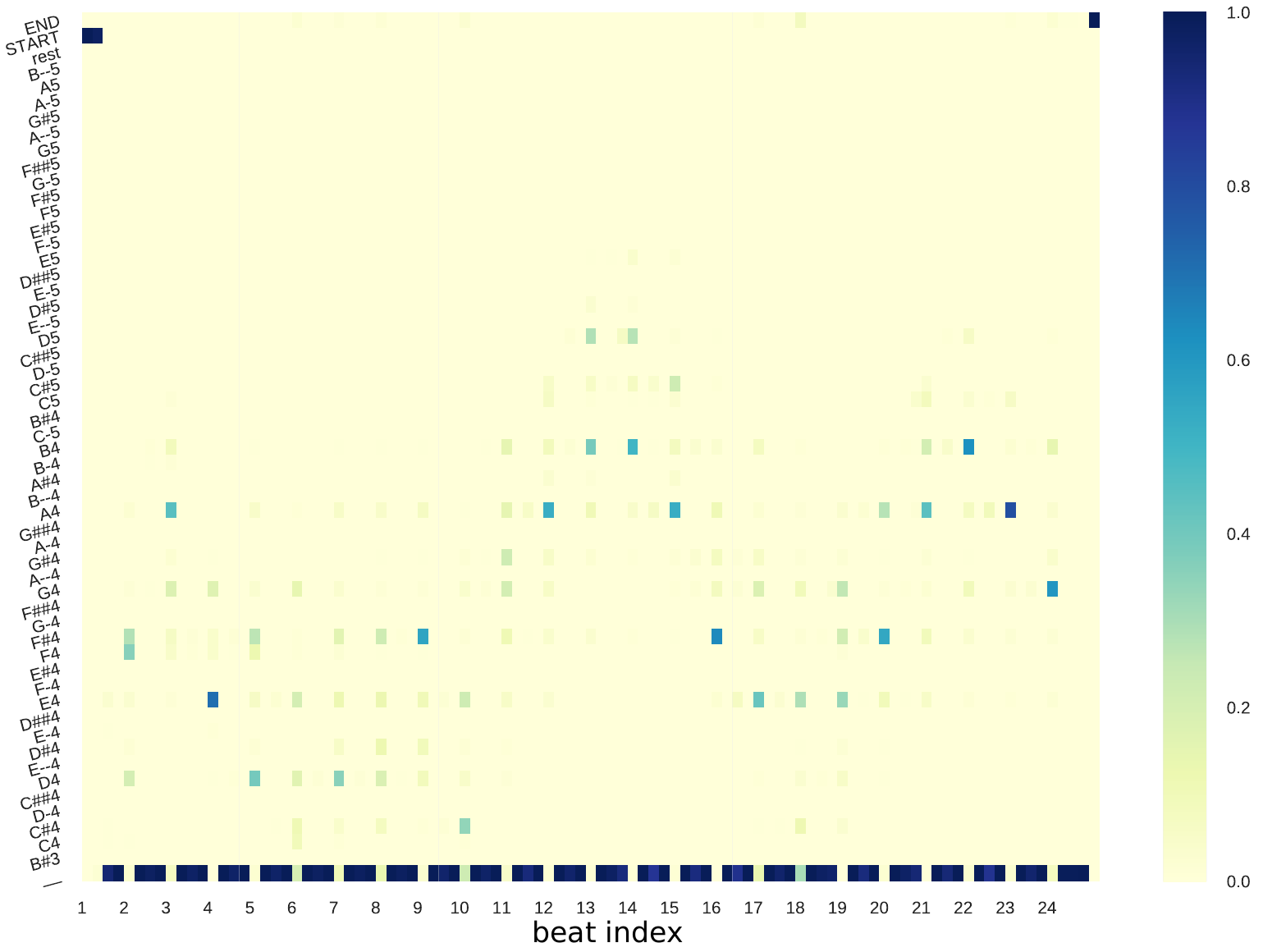}}
     \caption{Plot of $p(s_t| s_{<t})$ as a function of $t$ during the generation of the melody displayed in Fig.~\ref{fig:ex} in the constrained and unconstrained cases.
     }
     \label{fig:predictions}
     \end{figure}
     
 \begin{figure}
   \centering   
   \includegraphics[scale=0.8]{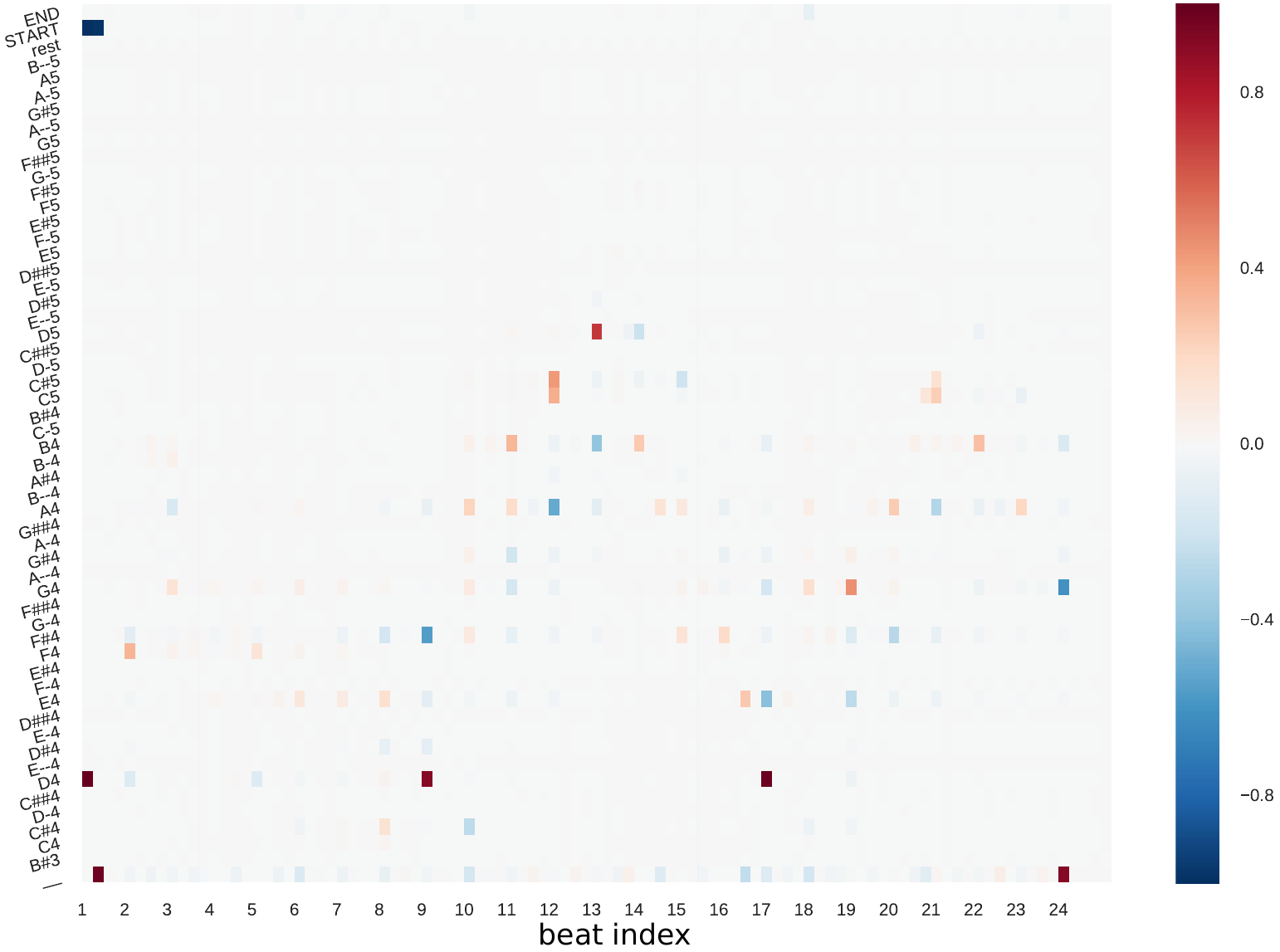}
 \caption{Difference between $p_{\text{constrained}}(s_t| s_{<t})$ and  $p_{\text{unconstrained}}(s_t| s_{<t})$ as a function of $t$ during the generation of the melody displayed in Fig.~\ref{fig:ex}.}
 \label{fig:diff}
\end{figure}

\label{sec:experimental-results}
We evaluated our architecture on the dataset of the melodies from the four-part chorale harmonizations by J.S. Bach. This dataset is available in the music21 Python package \cite{cuthbert2010music21}  and we extracted the soprano parts from all 402 chorales. In order to encode these monophonic sequences, we used the melodico-rhythmic encoding described in \cite{DBLP:conf/icml/HadjeresPN17}. The advantage with this encoding is that it allows to encode a monophonic musical sequence using only one sequence of tokens. This consists in adding an additional token ``$\texttt{\_\_}$'' which indicates that the current note is held. Furthermore, we do not use the traditional MIDI pitch encoding but used the real note names: among other benefits, this allows to generate music sheets  which are immediately readable and understandable by a musician and with no spelling mistakes. Time is quantized using a sixteenth note as the smallest subdivision (each beat is divided into four equal parts). An example of an encoded melody using this encoding is displayed in Fig.~\ref{fig:encoding}. We also perform data augmentation by transposing all sequences in all possible keys as long as the transposed sequence lies within the original voice range.

We used a 2-layer stacked LSTM \cite{hochreiter1997long} for both the Constraint-RNN and the Token-RNN using the PyTorch \cite{pytorch} deep learning framework and added a $20\%$ dropout on the input of the Token-RNN. Sequences are padded with $\mathtt{START}$ and $\mathtt{END}$ symbols.

Fig.~\ref{fig:ex1} shows  examples of the enforcement and the propagation of the constraints: even if generation is done from left to right, the model is able to generate compelling musical phrases while enforcing the constraints. In particular, we see that the model is able to ``anticipate'' the moment when it has to ``go'' from a low-pitched note to a high-pitched one and vice versa. The use of the melodico-rhythmic encoding allows to only impose that a note should be played at a given time, without specifying its rhythm. It is interesting to note that such a wide melodic contour (going from a D4 to a D5 and then going back to a D4 in only two bars) is unusual for a chorale melody. Nonetheless, the proposed model is able to generate a convincing Bach-like chorale melody.

    We now check how the constraints propagate backwards in time and how the constrained model deviates from the unconstrained model. For this, we compare the constrained model $p_{\text{constrained}}$ on the same  set of constraints  as in Fig.~\ref{fig:ex1} with its unconstrained counterpart  $p_{\text{unconstrained}}$. The latter is obtained by conditioning the model of Fig.~\ref{fig:archi} on a sequence of constraints in the special case where no constraint is set: the sequence of constraints is $(\mathtt{NC}, \dots, \mathtt{NC})$. Figure~\ref{fig:predictions} shows the evolution of  $p_{\text{constrained}}(s_t| s_{<t})$ and   $p_{\text{unconstrained}}(s_t| s_{<t})$ during the generation of the example in Fig.~\ref{fig:ex}. It is interesting to note that the conditional probability distributions returned by $p_{\text{constrained}}(s_t| s_{<t})$ are more concentrated on specific values than the ones returned by $p_{\text{unconstrained}}(s_t| s_{<t})$. The concentration of the all probability mass of $p_{\text{constrained}}(s_t| s_{<t})$ on  constrained notes confirms, on this specific example, that the proposed architecture has learned to enforce hard positional constraints. This assertion is experimentally verified on all  constrained sequences we generated.

    We also display in Fig.~\ref{fig:diff} the difference between the two distributions of Fig.~\ref{fig:predictions} for each time step. This highlights the fact that the probability mass distribution of $p_{\text{constrained}}$ is ``shifted upwards'' when the next positional constraint is higher than the current note, and ``downwards'' in the opposite case.

\begin{figure}
  \centering
  \includegraphics[width=0.7\textwidth]{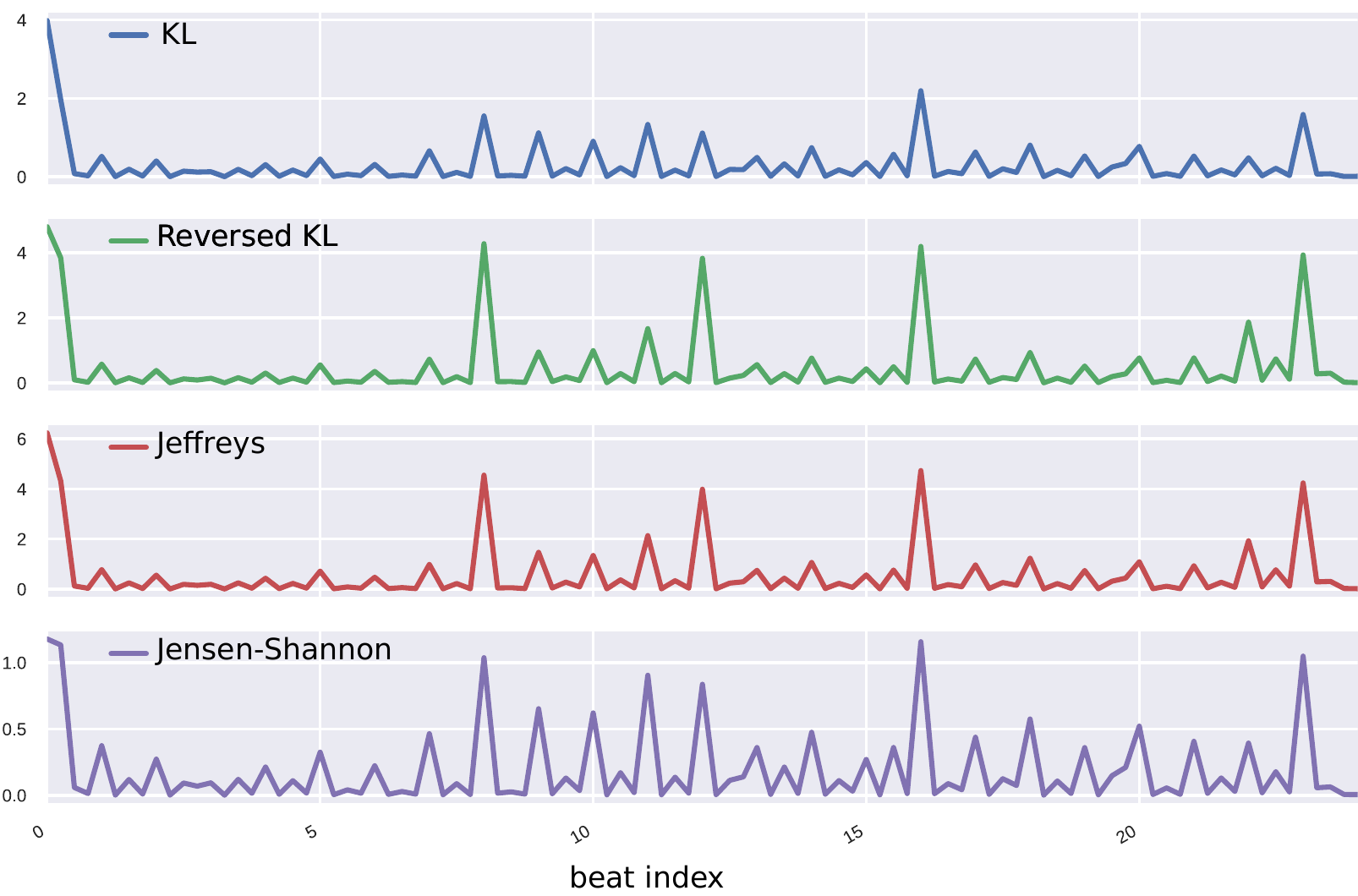}
    \caption{Square root of the divergence $D(p_{\text{unconstrained}}(s_t |s_{i<t})  || p_{\text{constrained}}(s_t |s_{i<t}))$ for the Kullback-Leibler, reversed Kullback-Leibler, Jeffreys and Jensen-Shannon divergences during the left-to-right generation of the example shown in Fig.~\ref{fig:ex}. The highest peaks correspond to the user-defined constraints (particularly clear when using the reversed Kullback-Leibler divergence) while the smaller ones demonstrate how the constraints tweaked the probability distributions in comparison with the unconstrained model.
      }
  \label{fig:kl1}
\end{figure}

\begin{figure}
  \centering
  \includegraphics[scale=0.7]{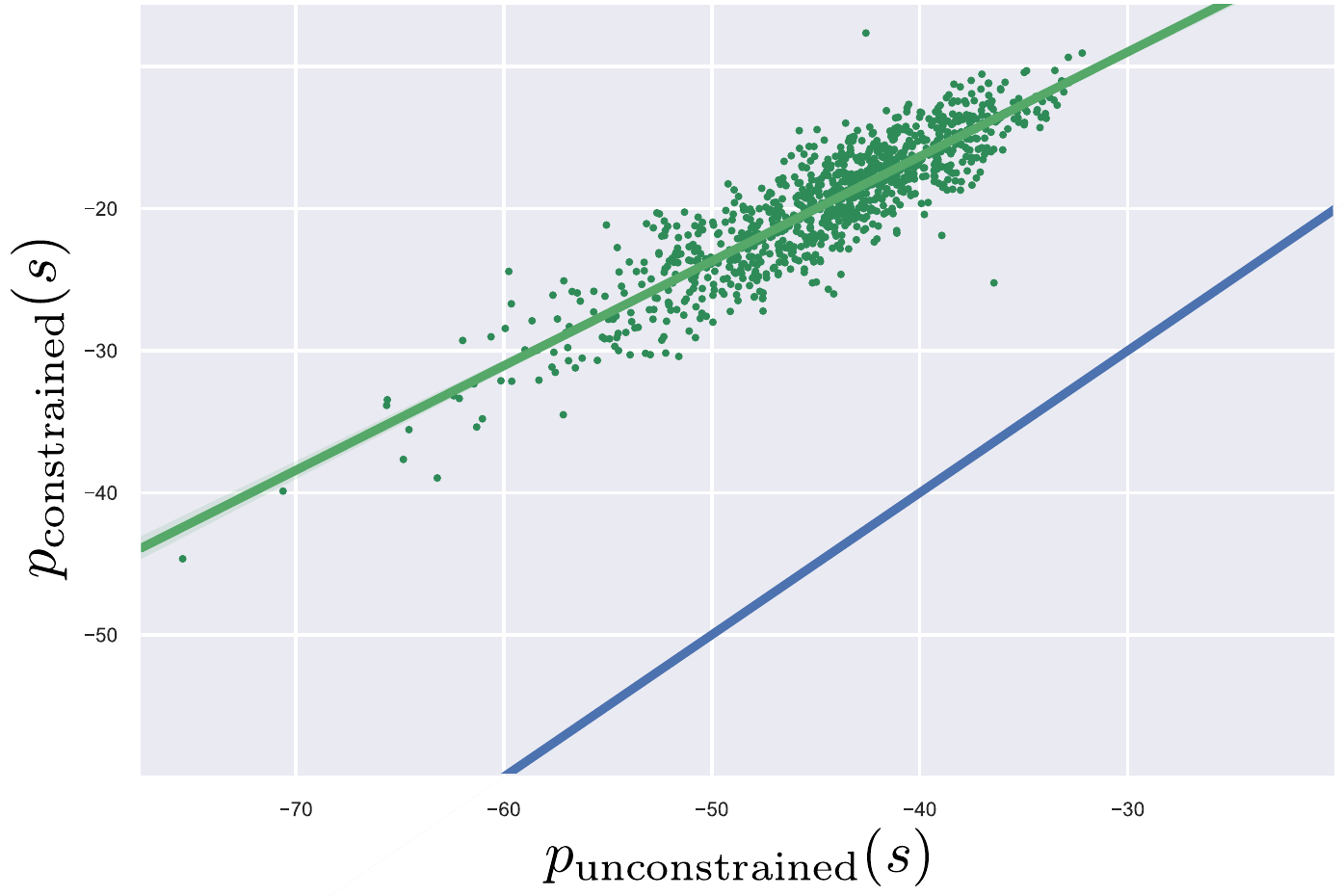}
    \caption{Point plot of $p_{\text{constrained}}(s)$ (y-axis) versus $p_{\text{unconstrained}}(s)$ (x-axis) on a set of 10000 generated (using $p_{\text{constrained}}$) sequences of length 96 (6 bars). The set of constraints is the one used in for the generations in Fig.~\ref{fig:ex1}. A logarithmic scale is used. The identity map is displayed in blue and the linear regression of the data points in green. The lines are closed to being parallel indicating the proportionality between
the two distributions, as desired.
      }
  \label{fig:ratio}
\end{figure}

We can quantify how the probability distributions $p_{\text{constrained}}(s_t| s_{<t})$ differ from  $p_{\text{unconstrained}}(s_t| s_{<t})$ by computing how dissimilar they are. In Fig.~\ref{fig:kl1} we plot the evolution of the square root of their divergence \cite{amari2007methods}:
\begin{equation}
  \label{eq:6}
D(p_{\text{constrained}}(s_t| s_{<t}) || p_{\text{unconstrained}}(s_t| s_{<t}))  
\end{equation}
 for different divergences. The divergences \cite{BR-2011} we considered are:
\begin{itemize}
\item the Kullback-Leibler divergence $D_{\text{KL}}(p || q) = \sum_i p_i \log\left( \frac{p_i}{q_i}\right)$,
  
\item the reversed Kullback-Leibler divergence $D_{\text{reversed KL}}(p || q) = D_{\text{KL}}(q || p)$,
  
\item the Jeffreys divergence  $D_{\text{Jeffreys}}(p || q) = D_{\text{KL}}(p || q) + D_{\text{KL}}(q || p)$,
\item the (symmetric) Jensen-Shannon divergence $D_{\text{JS}}(p || q) = \frac{1}{2} D_{\text{KL}}(p || m) +  \frac{1}{2} D_{\text{KL}}(q || m) $, where $m = \frac{p + q}{2}$.
\end{itemize}
This plot indicates  how the constraints are propagated backwards in time. The oscillation between high values of the divergences and the zero value is due to the encoding we chose as well as to the singularity of the musical data we considered. As can be seen in Fig.~\ref{fig:predictions}, the ``\texttt{\_\_}'' symbol concentrates most of the probability mass one time out of two since the soprano parts in Bach chorales are mostly composed of half notes, quarter notes and eighth notes. This is independent of the presence or absence of constraints so the constrained and unconstrained models make similar predictions on these time steps.

We now evaluate that the sampling using $p_\text{constrained}$ fulfills the requirements \eqref{eq:req1} and \eqref{eq:req2}. For a given set of constraints $\mathcal{C}$, we generated 10000 sequences and verified that the requirement \eqref{eq:req1} is fulfilled for all of these sequences (all constraints are enforced). In order to check the fulfillment of the requirement \eqref{eq:req2}, we plot for each sequence $s$ its probability in the constrained model $p_\text{constrained}(s)$ (defined as in Eq.~\eqref{eq:2}) as a function of $p_\text{unconstrained}(s)$ in logarithmic space. The resulting plot is shown in Fig.~\ref{fig:ratio}. The translation in logarithmic space indicates the proportionality between the two distributions as desired.

\section{Conclusion}
\label{sec:conclusion}
We presented the Anticipation-RNN, a simple but efficient way to
generate sequences in a learned style while enforcing positional
constraints. This method is general and can be used to improve many
existing RNN-based generative models. Contrary to other approaches, we
teach the model to learn to enforce hard constraints at training
time. We believe that this approach is a first step towards the
generation of musical sequences subjected to more complex constraints.

The constrained generation procedure is fast since it requires only $2N$ RNN calls, where $N$ is the length of the generated sequence; as it does not require extensive computational resources and provides an interesting user-machine interaction, we think that this architecture paves the way to the development of creative real-time composition software. We also think that this fast sampling could be used jointly with MCMC methods in order to provide fast initializations.

Future work will aim at studying how to improve the training of the model by carefully choosing the amount of masked notes (similarly to what is addressed in \cite{graves2017automated}), handling other types of constraints (imposing the rhythm of the sequences, enforcing the equality between two notes or introducing soft constraints) and developing responsive user interfaces so that the possibilities offered by this architecture can be used by a wide audience.

\bibliographystyle{abbrv}
\bibliography{constraints}

\end{document}